\documentclass[conference]{IEEEtran}
\IEEEoverridecommandlockouts
\usepackage{cite}
\usepackage{amsmath,amssymb,amsfonts}
\usepackage{algorithmic}
\usepackage{graphicx}
\usepackage{textcomp}
\usepackage{xcolor}
\def\BibTeX{{\rm B\kern-.05em{\sc i\kern-.025em b}\kern-.08em
    T\kern-.1667em\lower.7ex\hbox{E}\kern-.125emX}}

\begin{document}

\title{Subjective Assessment of Text Complexity: A Dataset for German Language}

\author{\IEEEauthorblockN{Babak Naderi\IEEEauthorrefmark{1},
Salar Mohtaj\IEEEauthorrefmark{1}, Kaspar Ensikat\IEEEauthorrefmark{1} and
Sebastian M\"oller\IEEEauthorrefmark{1}\IEEEauthorrefmark{2}}
\IEEEauthorblockA{\IEEEauthorrefmark{1}Quality and Usability Lab, Technische Universität Berlin, Berlin, Germany \\
\IEEEauthorrefmark{2}DFKI Projektbüro Berlin\\
\{babak.naderi $\mid$ salar.mohtaj $\mid$ sebastian.moeller\} @tu-berlin.de \\
\{ensikat\} @campus.tu-berlin.de}}

\maketitle
\begin{abstract}
This paper presents TextComplexityDE\footnote{Temporal URL: http://tiny.cc/mq643y}, a dataset consisting of 1000 sentences in German language taken from 23 Wikipedia articles in 3 different article-genres to be used for developing text-complexity predictor models and automatic text simplification in German language. The dataset includes subjective assessment of different text-complexity aspects provided by German learners in level A and B. In addition, it contains manual simplification of 250 of those sentences provided by native speakers and subjective assessment of the simplified sentences by participants from the target group. The subjective ratings were collected using both laboratory studies and crowdsourcing approach.
\end{abstract}

\begin{IEEEkeywords}
Text Complexity, Subjective Assessment, Dataset
\end{IEEEkeywords}
\section{Introduction}
Text is a major medium for transforming information in daily human to human communication.
Individuals with different backgrounds face various challenges when comprehending texts written in a more complex style \cite{saggion2017automatic}. 
Text complexity is defined as a metric that determines how challenging a text is for a reader \cite{common2010common}.
It influences the task load and consequently the Quality of Experience (QoE).
It has been shown in the domain of micro-task crowd working that the complexity of a task's description and instruction influences the workers' expected workload and consequently affect their decision on whether performing the micro-task or not \cite{naderi2018motivation}. 

Research in text complexity assessment and simplification dates back to the late 1940s. Since then, interest in such systems has grown especially in the last two decades Linguists developed guidelines for clear writing.
Researchers attempt to identify text complexity to determine  whether 1) a text needs simplification and 2) the text is suitable for a target group \cite{hancke2012readability}. Furthermore, automatic text simplification became an important area in Natural Language Processing.


Different factors can influence the complexity of text for readers.
From the lexical perspective, use of infrequent and non-familiar words, technical terminology and abstract concepts tend to increase the difficulty of the text \cite{temnikova2012text}.
Readers tend to struggle with issues at the syntactic level, such as long sentences and convoluted syntax which tend to cause processing difficulties \cite{harley2013psychology}.
Readers often struggle with ambiguous expressions and constructions which can arise at both lexical and syntactical levels. 
Those issues apply for all readers, some of may have a stronger effect on non-native speakers like grammatical attributes.


For German language, there are two guidelines for simplifying text for two different target groups. 
The \textit{Einfache Sprache} (easy language) is a convention fortargeting readers with weaknesses in reading and writing or those learning German as a second language.
The \textit{Leichte Sprache} (plain language) is another convention specifically designed for those with learning and comprehension disabilities. 

Text readability is described as the sum of all text elements that affect the reader’s understanding, reading speed and level of interest in the material \cite{dale1948formula}\footnote{This definition is very close to the definition of text complexity. As the later is a highly disputed term in linguistics we consider both to be synonym in this paper. For detailed discussion on text complexity see \cite{vulanovic2007measuring}}. 
The readability score is mostly measured using quantitative features.
By the 1980s, there were about 200 formulas.
The Flesch Reading Ease Score
, the Flesch-Kincaid readability formula and the Gunning Fog Index are the most prevalent formulas which are used to this day. 
They all use measures of average sentence length and average syllables per word for calculations.
However, existing formulas vary to a strong degree in their scores even when applied to the same material \cite{mailloux1995reliable}.
Furthermore, they are neither successful in sentence level assessment nor robust to unintelligible or meaningless sentences.

\subsubsection{Existing datasets}
Several text corpora are already collected with readability assessment and simplification in mind for some languages. 
However, most corpora focus only on article level i.e. either one readability score assigned to the entire article (e.g. \cite{schwarm2005reading} for English containing 2500 articles), or articles are classified to normal or simple  (e.g. PWKP data set containing Wikipedia articles and their corresponding Simple Wikipedia article).
 
For the German language, Klaper et al.\cite{klaper2013building} collected a parallel corpus of German text for normal and plain German by extracting articles from five websites which offer articles in both original and plain language. Similarly, Hancke et al. \cite{hancke2012readability} collected articles from two websites, one with the original text and the other article with the same topic but written for teenager audience. In both cases, text complexity is only differentiated between two levels. 
Others also used indirect measurements techniques like using eye-tracking, context questions, or measurements of effort to estimate the text complexity \cite{Barrieren}.


The rest of this paper is organized as follows. In the next section, we explain the dataset structure and how we collected and evaluated the ratings.
In Section III, we describe the process of manual simplification and finally in section IV, we discuss our findings and present implications for future works.

\section{Subjective Assessment of Text Complexity}
For this dataset, we decided to collect subjective ratings in the sentence level and focus on German learners as our target group. 
We adapted standards \cite{p800,p808} on how to perform Absolute Category Ratings in the study design and data screening process to collect reliable and valid data.

\subsection{Source Text}
Sentences contained in the dataset were collected from two sources; the online encyclopedia Wikipedia\footnote{de.wikipedia.org} for German (1000 sentences) and the Leichte Sprache (Simple language) dataset developed by Klaper et al. \cite{klaper2013building} (100 sentences).
Sentences from the Leichte Sprache were used as Gold Standard Questions \cite{naderi2018motivation} i.e. indicator for the quality of data collected in a rating session.
We took 23 articles from three domains in Wikipedia and two articles from the Leichte Sprache. In total, the dataset holds 1019 sentences from Wikipedia (and from history, society and science domains).

\subsection{Development of Scale}
We conducted a pilot study to determine relevant dimensions of text complexity that can be captured within the subjective assessment.
An initial item pool with 11 questions was developed and reviewed by a linguistic expert (cf. Table \ref{table:1}).
Within a pilot study, 100 sentences were assessed by crowd workers. We created 20 crowdsourcing jobs in the ClickWorker micro-task crowdsourcing platform. 
In each job, participants assessed five sentences (4 from Wikipedia pool and one from Leichte Sprache) by answering the 11 questions from the item pool.
In total, ten different workers rated each sentence.
Overall, 77 German learners participated in the study (submission from native speakers were discarded).
\begin{footnotesize}
\begin{table}
\centering
\caption{Initial item set used in the pilot study.}
\begin{tabular}{|l p{0.4\textwidth} |} 
 \hline
  & Item \\
 \hline
 1 & How do you rate the overall complexity of the sentence? \\ 
 2 & How difficult was it for you to read this sentence? \\ 
 3 & How familiar are you with the topic of the article? \\ 
 4 & How difficult would it be to translate this sentence into your native language?\\ 
 5 & How many different ways can this sentence be interpreted?\\ 
 6 & How difficult would it be to explain this sentence to another person?\\ 
 7 & How well did you understand the sentence? \\ 
 8 & How many words in this sentence are unfamiliar to you?\\ 
 9 & Take a look at the hardest words contained in the sentence. How difficult is it for you to understand those words?\\ 
 10 & How many words in this sentence have multiple interpretations?\\ 
 11 & How do you rate the complexity of the syntactical structure of the sentence?\\  
 \hline
\end{tabular}

\label{table:1}
\end{table}
\end{footnotesize}
\subsubsection{Data Screening}
\label{section:ds}
First, submissions from workers with unreasonable completion times or unrealistic answer to the gold standard question were removed. 
Next, responses were evaluated against unexpected patterns in ratings (i.e. no variance or potential outliers). 
Univariate outliers were identified in item level by calculating the standardized scores (absolute z-score larger than 3.29 considered to be a potential outlier \cite{naderi2018motivation}).
Submissions with more than one potential outliers were removed. Finally, 122 answer packages (i.e. 610 ratings) were accepted.

\subsubsection{Evaluation}
The Mean Opinion Score (MOS) value for each item was calculated per sentence using the accepted answer packages. 
Ten sentences were removed as there less than five votes for each were available.
Items were investigated by calculating the Cronbach's $\alpha$ value, the internal consistency, assuming that all items express the same construct i.e. text complexity. 
The $\alpha$ value of .996 is achieved by removing 4 items (i.e. item 3, 5, 8 and 10).
Next, Principal Component Analysis was performed which leads to extract two factors. 
Complexity (item 1) was the dominant item loading on the first factor and Understandability (item 7) was loading on the second factor. 

\subsection{Data Collection Procedure}
An online survey system was created to collect the subjective assessment of 1000 sentences using three items each rated on a 7-point Likert Scale.
A survey session consist of training and rating sections.
The training section was containing three sentences which participants needed to rate on the same scale as the main section.
The sentences in the training section were constant and  represent very easy, average and very complex sentences.
Afterward, participants rated complexity, understandability and lexical difficulty\footnote{This item was included as we aim
to investigate it in future.} of ten sentences by answering to the following questions on 7-point Likert scales:

\textbf{Complexity:} \textit{How do you rate the complexity of the sentence?} 
Scale from \textit{very easy (1)} to \textit{very complex (7)}.

\textbf{Understandability:} \textit{How well were you able to understand the sentence?} 
Scale from \textit{fully understood (1)} to \textit{didn’t understand at all (7)}.

\textbf{Lexical difficulty:} \textit{Regarding the hardest words in the sentence: How difficult is it to you, to
understand these words?} 
Scale from \textit{very easy (1)} to \textit{very difficult (7)}.



Users could participate in the survey as many times as they wanted and the system was designed to avoid a same sentence to be assigned to the same participant on their return. 

\subsubsection{Participants}
We aimed to collect at least ten votes per each sentence.  Overall, 369 individuals participated in the study.
From them, 267 reported a German language level between A and B. In total, the survey was completed 1322 times. Out of those 1065 were provided valid ratings from German language learners resulting in 10650 valid sentence ratings split across the 1000 sentences.
Participants were recruited from three channels: 
Paid Crowdsourcing (16\% of valid answers), Volunteers\footnote{Through 87 Facebook groups organized by German learners.} (21\% of valid answers) and Laboratory study\footnote{33 German learners   laboratory sessions of 1 to 1:30 hours} (63\% of valid ratings).
Participant on average were 32 years old ($StD=8.9$) and mostly reported to hold a university degree (64\%).

\subsubsection{Absolute Category Ratings}
\begin{figure}[htbp]
	\centerline{\includegraphics[width=0.45\textwidth]{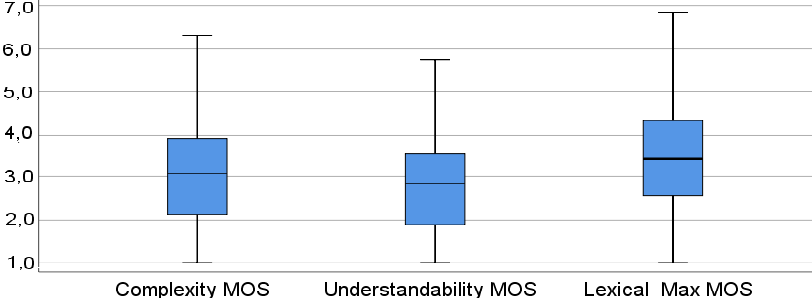}}
	\caption{Distribution of MOS of Complexity, Understandability and Lexical difficulty ($N=1000$)}
	\label{box}
\end{figure}

Following the data screening process explained in Section~\ref{section:ds}, 5 to 18 valid ratings for each sentence remained in the dataset. For each sentence MOS, standard deviation and 95\% Confidence Intervals of each dimension are reported in the dataset.
Fig.~\ref{box} illustrates the distribution of MOS values.
As expected sentences from Wikipedia are more complex ($M_{MOS}=3.22$) than the sentences from the plain language dataset ($M_{MOS}=1.2$). 
In addition there are strong significant correlations between the three dimensions: complexity has a correlation of .896 with understandability and .905 with lexical difficulty. Understandability has a correlation of .935 with lexical difficulty. 

In addition we used Amstad's adaptation of Flesch-Reading-Ease (FRE) \cite{amstad1978verstandlich} formula to calculate the readability score for both article (n=25) and sentence level.
We used the collected ratings to calculate complexity score in article level\footnote{We used average as a very basic model.}. The FRE-scores for the two articles written in plain language were 62 and 66 (out of 100) interpreted as moderately difficult, while participants in our study considered them to be very easy.
The FRE-Score and average MOS rating of complexity in article level strongly correlate ($r = .89$, $p<.001$) but with a huge intercept when they are normalized (2.6 in 7 point range). In the sentence level both values moderately correlate ($r = .55$, $p<.001$). Strong disagreement were observed in highest range of FRE-Score. This is in-line with previous research that the FRE-formula does not perform well at sentence-level \cite{mcclure1987readability}.
Finally, the sentences with highest complexity rating were examined by native speakers. 
It revealed that, they are either thematically complex even for the native speakers or are written in a convoluted manner.

\section{Manual Simplification}
265 sentences with complexity rating above 4 point of MOS and understandability rating above 3.5 MOS were selected for manual simplification. 
Overall, 659 simplifications of original sentences were collected from 75 native speakers. For 250 out of 265 sentences at least one simplification were provided. In 90 cases native speakers reported that they were unable to simplify the provided sentence. 

\section{Discussion and Future Work}

This work presents a corpus of 1000 sentences in German language which their complexity, understandability and lexical difficulty were assessed by group of language learners participated in subjective studies conducted following best practices in the quality of experience community. 
In addition, the dataset contains manual simplifications for 250 of those sentences written by native speakers.
It should be noted that subjective ratings refer to the degree that participants perceived a concept. For some aspect of text it might be important to not only measure the perceived degree but also the actual value of concept. For instance, the understandability measurement in this study refers to how good participants think they understood the given text. It may differ from the actual level of understanding for which different assessment methods like content questions should be used. Therefore, researchers should carefully decide which kind of measurement technique to employ depending to the goals of their study. 
For future work, we would like to compare subjective assessment of text understandability and complexity as explained in this paper with actual understandability (e.g. measured by content questions) and readability (e.g. measured by eye-tracking) scores.  



\bibliographystyle{IEEEtran}
\bibliography{paper}

\end{document}